\documentclass[letterpaper, 10 pt, conference]{ieeeconf}

\IEEEoverridecommandlockouts                              % This command is only needed if 
\usepackage{amsmath} % assumes amsmath package installed
\usepackage{amssymb}  % assumes amsmath package installed
\usepackage[dvipsnames]{xcolor}
\usepackage{graphicx}
\usepackage{url}
\usepackage{booktabs}
\usepackage{subcaption}
\usepackage{algorithm}
\usepackage{algorithmicx}
\usepackage{algpseudocode}
\usepackage{comment}
\algrenewcommand\algorithmicrequire{\textbf{Input:}}
\algrenewcommand\algorithmicensure{\textbf{Output:}}
\usepackage{geometry}
\newcommand\Ss{\mathbb{S}}

\newcommand\QQ{\mathcal{Q}}
\newcommand\qq{\mathbf{q}}
\newcommand\DD{\mathcal{D}}
\newcommand\dd{\mathbf{d}}

\newcommand\GG{\mathcal{G}}

\newcommand\OO{\mathcal{O}}

\newcommand\EE{\mathcal{E}}
\newcommand\ee{\mathbf{e}}
\newcommand\RR{\mathbb{R}}
\newcommand\xx{\mathbf{x}}
\newcommand\nn{\mathbf{n}}

\newcommand\JJ{\boldsymbol{J}}
\newcommand\RVDSA{\text{RVDSA}}

\title{\LARGE \bf
Anatomy-Aware Dexterity-Driven Design Optimization of Surgical Continuum Robots
}

\author{Tony Qin$^{1}$, Peter Connor$^{2}$, Khoa Dang$^{3}$, Carter Hatch$^{3}$, Caleb Rucker$^{3}$, Robert J. Webster III$^{2}$,\\
and Ron Alterovitz$^{1}$% <-this % stops a space
\thanks{*This work was supported in part by the National Institutes of Health (NIH) under award R01EB032385 and the National Science Foundation (NSF) under award 2008475. The content is solely the responsibility of the authors and does not necessarily represent the official views of NIH or NSF.}% <-this % stops a space
\thanks{$^{1}$Tony Qin and Ron Alterovitz are with the Department of Computer Science, University of North Carolina at Chapel Hill
        {\tt\small tonyqin@cs.unc.edu}}%
\thanks{$^{2}$Peter Connor and Robert J. Webster III are with the Department of Mechanical Engineering, Vanderbilt University.%
}
\thanks{$^{3}$Khoa Dang, Carter Hatch, and Caleb Rucker are with the Department of Mechanical, Aerospace, and Biomedical Engineering at The University of Tennessee, Knoxville.%
}
}
\begin{document}

\maketitle
\thispagestyle{empty}
\pagestyle{empty}

%%%%%%%%%%%%%%%%%%%%%%%%%%%%%%%%%%%%%%%%%%%%%%%%%%%%%%%%%%%%%%%%%%%%%%%%%%%%%%%%
\begin{abstract}
Performing complex medical procedures with continuum robots requires careful selection of their geometric design parameters. The robot should have high dexterity in the specific anatomical environment of its procedure. This work presents a design optimization method that considers both dexterity and anatomy. We introduce the Reachable Volumetric Dexterous Solid Angle (RVDSA) metric as our objective, which measures the ability of a robot's end effector to reach the points in a goal volume from different directions via collision-free paths from a start configuration. We present a computationally efficient motion planner to compute this objective function for a given robotic design, and we use an asymptotically optimal simulated annealing optimizer to compute an optimized design. We applied our new method to optimize the design of a bimanual dexterous sheaths robot for performing procedures on cancerous polyps in colon anatomies, achieving a 78\% higher RVDSA on average than optimizing for 3D voxel coverage alone.
\end{abstract}

\section{INTRODUCTION}
Surgical continuum robots are well suited for minimally invasive surgical procedures since their flexible, curvilinear bodies allow them to operate within highly constrained workspaces in the body. There are several different classes of continuum robots, and each type of robot has design parameters that can be chosen to tailor the robot for a specific procedure~\cite{Dupont2022_Continuum}. The design optimization for surgical continuum robots can consider the criteria of the anatomical environment \cite{Bedell2011_Design, Torres2012_Taskoriented} and the dexterity of the robot \cite{Leibrandt2023_Designing}.

Anatomy-informed design optimization of surgical continuum robots enables them to be tailored for a specific procedure or individual anatomy. One of the most basic requirements is for the robot to be able to reach a goal in the anatomy while avoiding obstacles \cite{Baykal2017_Asymptotically}. These metrics typically measure reachability of a goal point $g \in \RR ^3$ or volume $\GG \subset \mathcal{W} \subset \RR ^3$ in the workspace $\mathcal{W}$.

\begin{figure}[!t]
    \centering
    \begin{subfigure}[t]{.49\linewidth}
        \includegraphics[width=\linewidth]{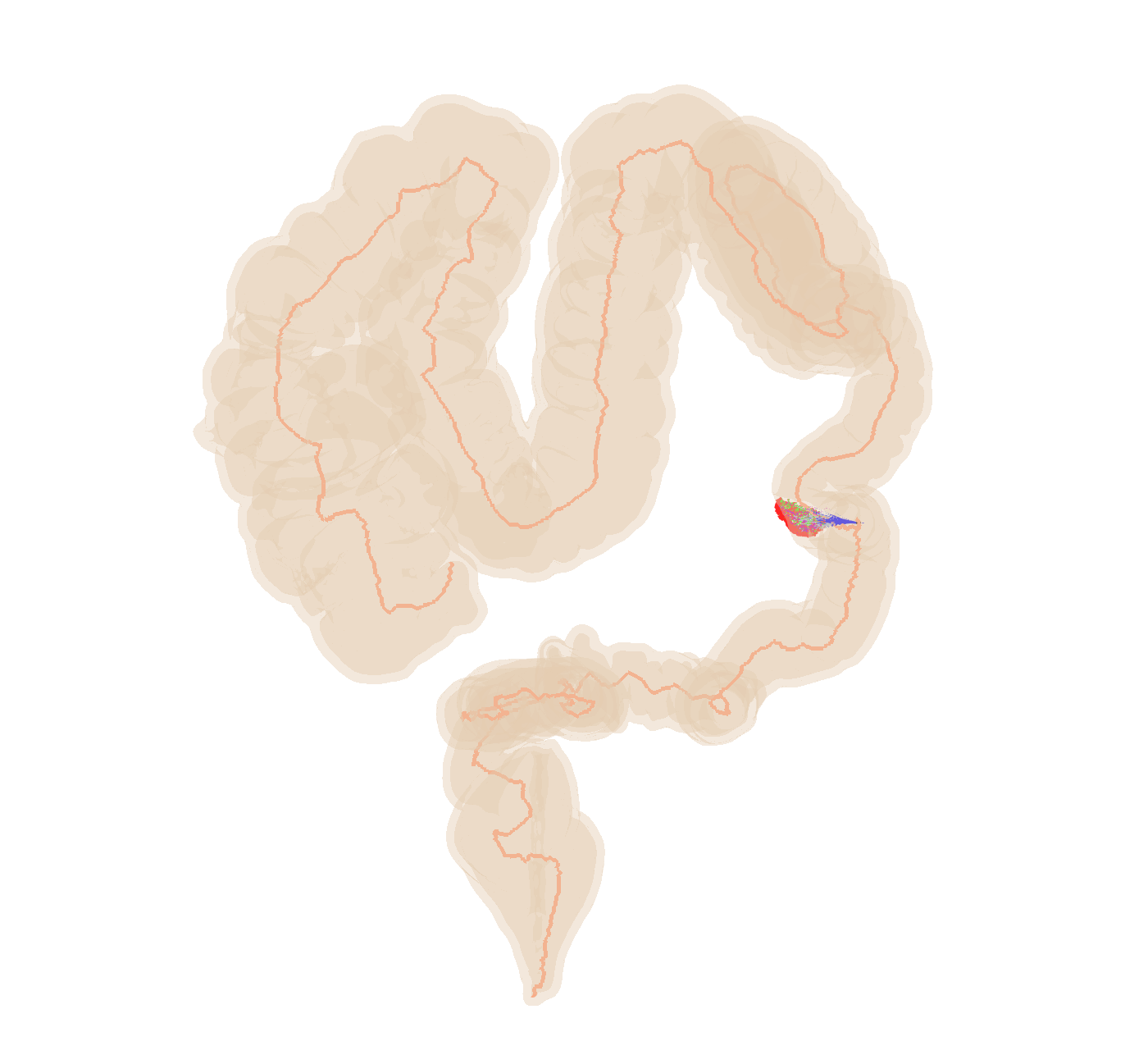}
        \caption{}
    \end{subfigure}
    \begin{subfigure}[t]{.49\linewidth}
        \includegraphics[width=\linewidth]{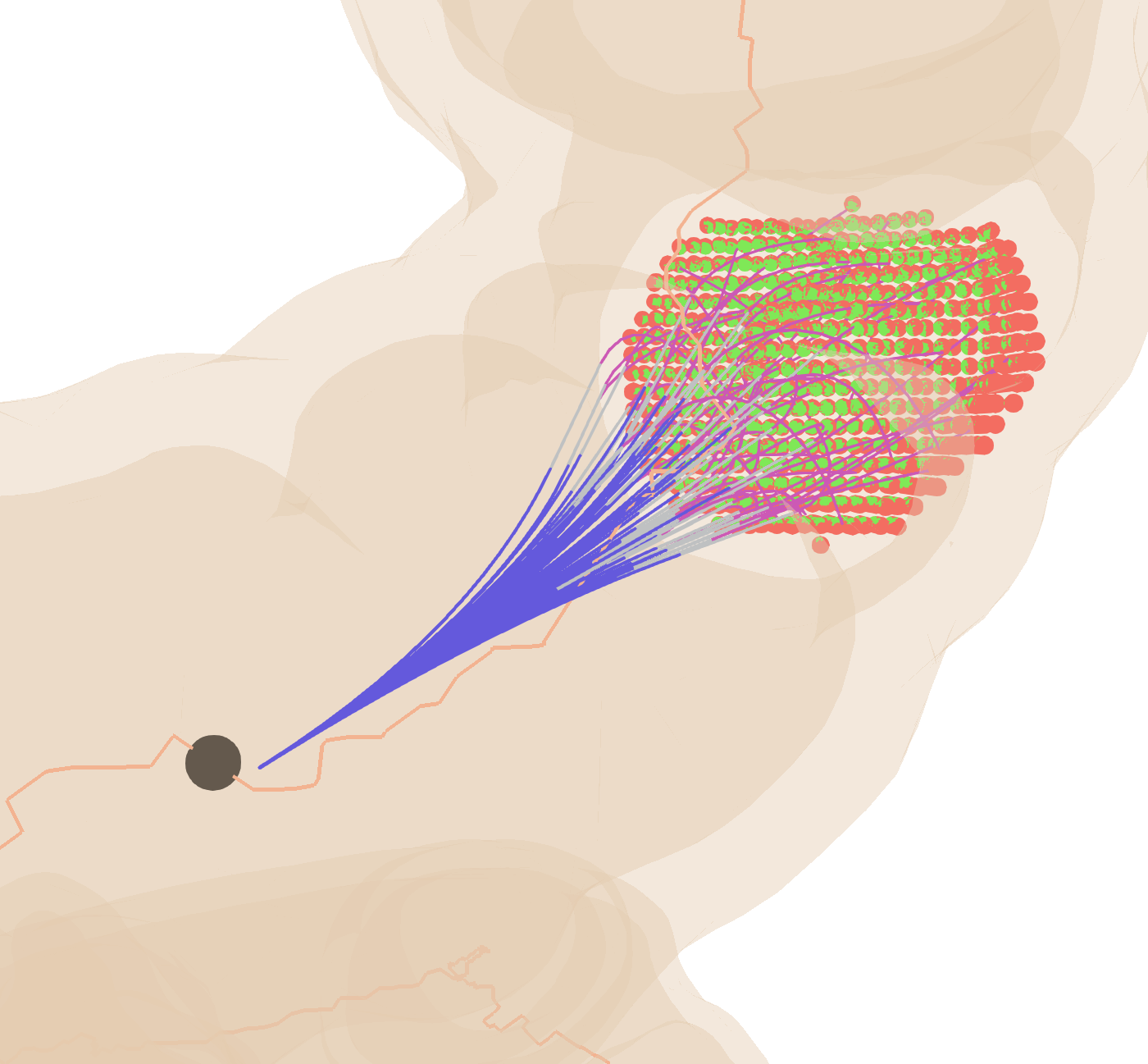}
        \caption{}
    \end{subfigure}
    \\[6pt]
    \begin{subfigure}[t]{.49\linewidth}
        \includegraphics[width=\linewidth]{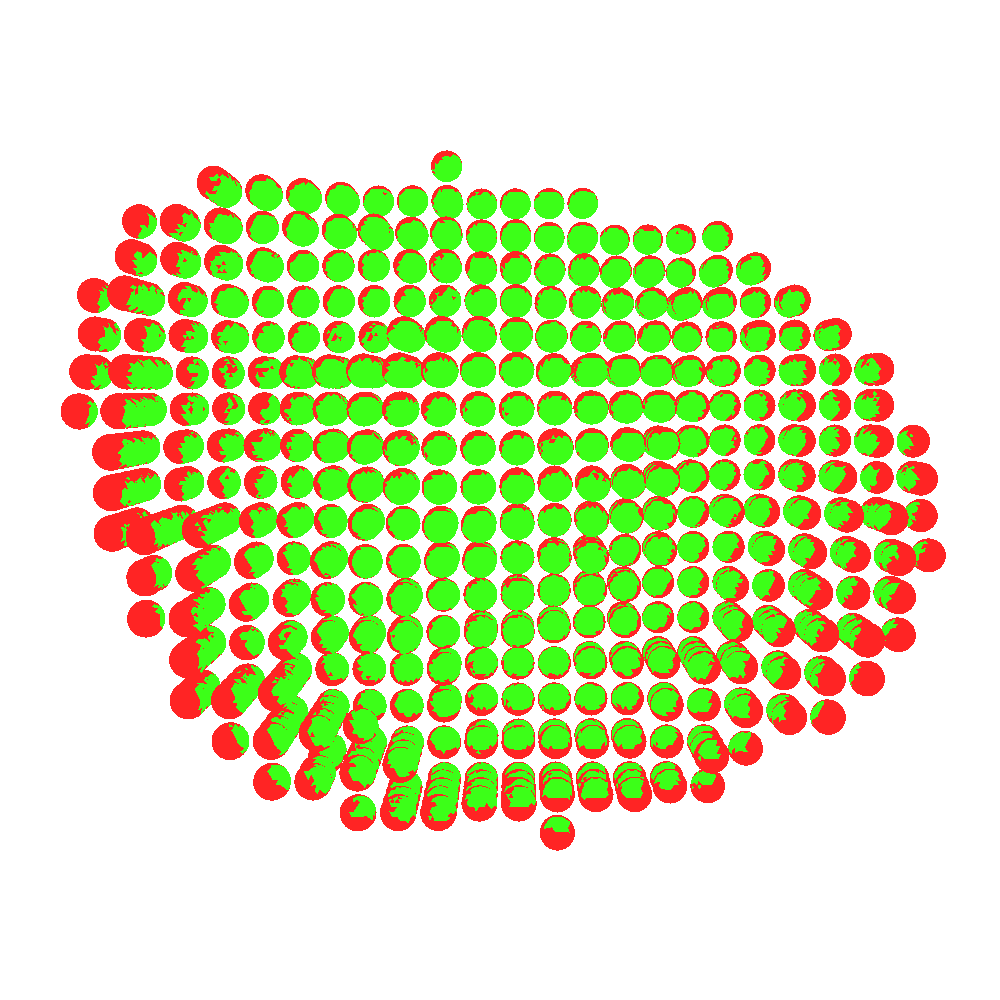}
        \caption{}
    \end{subfigure}
    \begin{subfigure}[t]{.49\linewidth}
        \includegraphics[width=\linewidth]{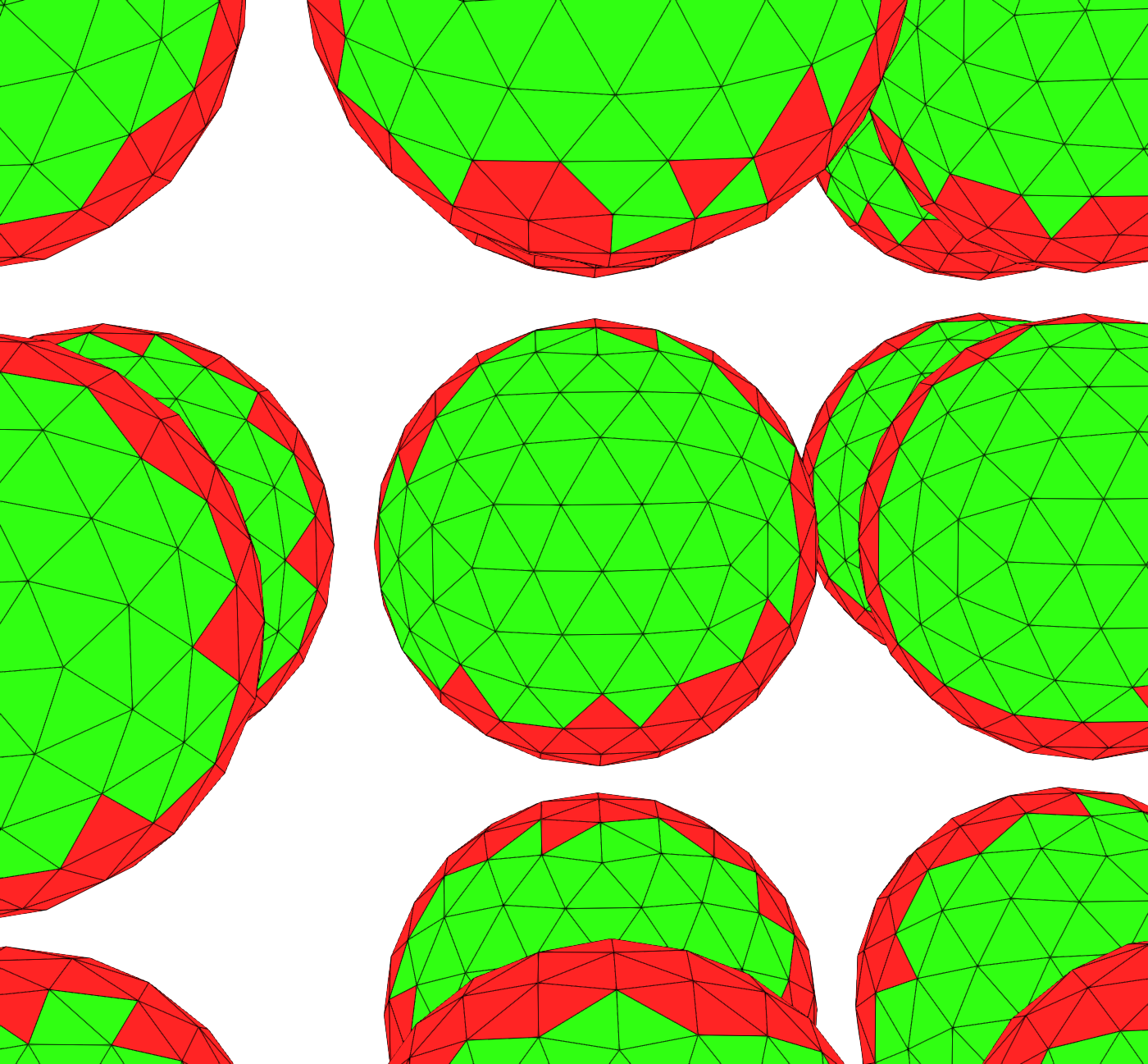}
        \caption{}
    \end{subfigure}
    
    \caption{We define the RVDSA metric to evaluate the quality of the geometric design of a surgical continuum robot and apply it to the dexterous sheaths robot for procedures in the colon. (a) Colon anatomy from a CT scan and computed centerline. (b) Constant curvature backbones reaching target voxels in the colon anatomy from the computed colonoscope location depicted as a black dot. (c) The discretization of the goal volume $\GG$ into voxels. Each polyp voxel has a discretized unit sphere representing the directional component of the end effector. (d) Discretization of the unit sphere $\Ss ^2$ into a subdivided icosahedron. Each triangle on the discretized unit sphere is marked green if it is reached at that voxel and direction. The RVDSA metric is computed from the percentage of triangles reached.}
    \label{fig:intro}
\end{figure}

On the other hand, dexterity metrics for surgical robots allow for the design of highly dexterous manipulators \cite{Wang2019_Task, Wang2021_Dexterity}. A classic dexterity metric is the Dexterous Solid Angle (DSA), which measures the range of directions from which a robot's end effector can reach a given point, expressed as the coverage of a unit sphere $\Ss ^2$~\cite{Abdel-Malek1994_Dexterous}. The DSA was extended to the entire reachable workspace by averaging the DSA across the workspace volume, effectively calculating a metric over $\mathcal{W} \times \Ss ^2 \subset \RR ^3 \times \Ss ^2$ \cite{Wu2017_Dexterity}.

When selecting design parameters for surgical continuum robots, there is typically a trade-off between the dexterity across the whole workspace and the dexterity at specific sites within the workspace, so maximizing dexterity specifically in regions of clinical interest can enhance their potential. A medical procedure may require high dexterity in certain regions of the workspace, such as near a cancerous polyp in the colon \cite{Connor2025_System}, a nodule in the lung \cite{Baykal2017_Asymptotically}, or a patent foramen ovale in the heart \cite{Gosline2012_Percutaneous}. Furthermore, the robot must operate within the constraints of the anatomical environment.

Minimally invasive procedures using continuum robots are often performed in narrow tubular environments with little room to navigate laterally. Therefore, an anatomically aware design optimization for dexterity will lead to a more suitable robot design for the specific procedure under consideration.

Another challenge is the computational cost of evaluating the dexterity of a robot. The standard approach is to use Monte Carlo sampling to evaluate the dexterity \cite{Leibrandt2023_Designing, Wang2019_Task}. While this approach is applicable for workspace analysis, it may be less efficient if only a subset of the workspace is relevant. An efficient evaluation of a dexterity metric is required to enable design optimization of complex surgical continuum robots, during which hundreds to thousands of designs may need to be evaluated.

In this paper, we introduce a computationally efficient method to optimize the kinematic design parameters of a surgical continuum robot to maximize its dexterity across goal volumes in anatomical environments. We extend the Dexterous Solid Angle to define the \textit{Reachable Volumetric Dexterous Solid Angle} (RVDSA) of a robotic manipulator design as the proportion of approach directions across the goal volume, $\GG \times \Ss ^2$, achievable by its end effector with a collision-free motion from a start configuration. We provide a mathematical definition of the RVDSA of a particular design, and we present a motion planner to efficiently compute a discretized version of the objective. We use a version of Adaptive Simulated Annealing that incorporates our motion planner in a manner that provides asymptotic optimality, guaranteeing convergence toward a globally optimal design as compute budget increases \cite{Baykal2017_Asymptotically}.

Our design optimization method can be applied to any surgical continuum robot scenario in which (1) a finite number of design parameters influence the robot's kinematic model and (2) maximizing dexterity at goal volumes in the anatomy is key for task success. We apply our design optimization method to the bimanual Dexterous Sheaths robot for Endoscopic Submucosal Dissection (ESD), a multi-stage procedure to remove large cancerous tumors in the colon.

The contributions of our paper are as follows:
\begin{enumerate}
    \item Combining both anatomical and dexterity considerations for design optimization of a surgical continuum robot. We present the Reachable Volumetric Dexterous Solid Angle (RVDSA), a measure of a robot's ability to dexterously reach a goal volume from many different directions, $\GG \times \Ss ^2$, from an initial configuration while avoiding collisions in the anatomy.
    \item An efficient RVDSA evaluator that adapts Jacobian-guided and self-motion sampling strategies within a sampling based motion planner.
    \item Application of our design optimization to the dexterous sheaths robot, maximizing the RVDSA on multiple colon anatomies. 
\end{enumerate}

\section{RELATED WORK}
\subsection{Dexterity Analysis}
A suitable metric for the dexterity of a surgical robot is important to evaluate possible kinematic designs. A traditional measure of dexterity is the manipulability ellipsoid, which measures an end effector's freedom of movement at a specific configuration \cite{Yoshikawa1985_Manipulability}. This idea has been extended with obstacle penalty terms \cite{Vahrenkamp2015_Representing} and applied to surgical continuum robots \cite{Leibrandt2017_Implicit}. However, these measures are difficult to apply globally to compare different robot designs or architectures.

A basic requirement of dexterity is that the end effector should be able to reach the workspace from many different orientations. The Dexterous Solid Angle (DSA) measures the proportion of directions from which a robot's end effector can reach a particular point. The methods using DSA discretize $\Ss ^2$, the unit sphere, and determine how much of this discretized sphere is reachable by a robot's end effector~\cite{Abdel-Malek1994_Dexterous, Wu2017_Dexterity}. While DSA was originally only considered for a single point, later work expanded this metric to the whole workspace \cite{Wu2017_Dexterity, Wei2024_Dexterity}, effectively computing the metric for a 5-volume in $\RR ^3 \times \Ss ^ 2$. While considering the entire workspace can measure the global dexterity of the robot, it is not tailored to the specific regions in the workspace that are of interest. 

Leibrandt et al. introduce a similar discretized dexterity metric in $SE(3)$. In addition, their method can restrict the dexterity calculation to an anatomically relevant subset of the workspace~\cite{Leibrandt2023_Designing}. However, their method uses a compute-intensive Monte Carlo sampling method to evaluate their dexterity metric, and they do not consider whether the end effector poses can be reached with a collision-free path in the environment. Additionally, randomly sampling configurations from the configuration space \cite{Wu2017_Dexterity, Wang2021_Dexterity, Leibrandt2023_Designing, Bamoriya2025_Data} is inefficient for computing the DSA for a restricted task space where only a small subset of the workspace is of interest since most of the randomly sampled robot configurations may not have its end effector in the task space.

\subsection{Motion Planning}
Sampling-based motion planners such as rapidly exploring random trees (RRTs) or probabilistic roadmaps (PRMs) are well suited to address the problem of finding collision-free paths. They compute collision-free paths in configuration space from a start configuration to a goal configuration by sampling collision-free configurations to add to their graph.

However, solving the inverse kinematics problem is challenging for continuum robots. There has been progress to provide closed-form analytical solutions to the inverse kinematics of constant curvature robots \cite{Garriga-Casanovas2018_Kinematics, Qiu2023_Efficient}, but a closed-form solution has not yet been found for complex, multi-segment robot architectures.

One method to solve the inverse kinematics problem is to use an end effector Jacobian-based controller \cite{Liegeois1977_Automatic, Wampler1986_Manipulator}. The method of Jacobian-based control was combined with sampling-based motion planners with the JT-RRT method, which uses both random sampling of the configuration space as well as sampling based on the Jacobian controller \cite{VandeWeghe2007_Randomized}. This speeds up a motion planner by biasing its exploration toward a goal in the workspace. This strategy was used to efficiently compute paths to target points in a workspace for continuum robots \cite{Meng2022_RRTBased}.

\subsection{Design Optimization}
Design optimization for surgical continuum robots has used a wide variety of derivative-free optimizers such as Pattern Search \cite{Bedell2011_Design}, Nelder-Mead \cite{Burgner2013_Computational}, Particle Swarm Optimization \cite{Wang2021_Dexterity}, and BOBYQA  \cite{Leibrandt2023_Designing}.

Baykal et al. introduce an asymptotically optimal optimization method for robotic designs using a motion planner, based on the Adaptive Simulated Annealing algorithm \cite{Baykal2019_Asymptotically}. It achieves this by using a sampling-based motion planner to evaluate an objective function, iteratively increasing the compute time for this motion planner to guarantee convergence to the optimal design.

\section{PROBLEM DEFINITION}
\subsection{Robot Design}
We aim to find the best design $\dd$ for our robot within the design space $\DD \subset \RR ^n$, where $n$ is the number of design variables. Each design $\dd$ specifies kinematic design parameters of the robot that are fixed throughout the procedure, while the configuration $\qq$ defines the values of the actuation variables that can change during a procedure. Let $Shape(\dd, \qq) \subset \RR ^3$ represent the set of points that the robot occupies in 3D space with a configuration $\qq$ under some design $\dd$.

\subsection{Environment}

We define a static environment $\ee$ in which the robot operates that contains empty space, an obstacle volume, and a goal volume. Let $\GG \subset \RR ^3$ and $\OO \subset \RR ^3$ represent the goal volume and the obstacle volume, respectively. We use finite a set of environments $\mathbf{E} = \{\ee _1, \dots , \ee _m\}$, where $m$ is the number of environments drawn from some environment space $\EE$.

A configuration is collision free in the environment if $Shape(\dd, \qq) \cap \OO = \emptyset$. We define a continuous function $\sigma : [0, 1] \rightarrow \QQ$ called a path where $\sigma(0) = \qq_0$, a start configuration that we assume to be the zero vector $\qq_0 = \mathbf{0}$. We define a collision-free path as $\sigma_{\text{free}} ^{\dd, \ee}: [0, 1] \rightarrow \QQ$ if it satisfies $\forall \tau \in [0, 1], Shape(\dd, \sigma_{\text{free}} ^{\dd, \ee} (\tau )) \cap \OO = \emptyset$.

\begin{figure}
    \centering
    \includegraphics[width=\dimexpr\columnwidth]{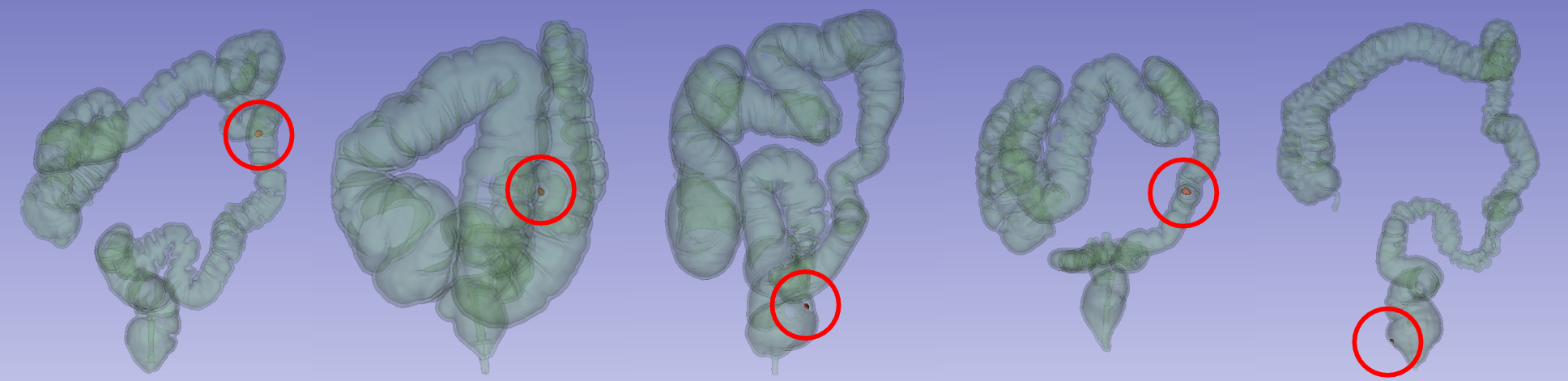}
    \caption{The five colon anatomies of our dataset with the location of each polyp circled in red.}
    \label{fig:dataset}
\end{figure}
\subsection{Objective}

We wish to maximize the RVDSA of our robot on our set of environments, $\mathbf{E}$. We consider the 3D location that the end effector reaches as well as the approach vector at that location. This is the 5-volume $\GG \times \Ss ^2  \subset \RR ^3 \times \Ss ^2$, where $\Ss ^2$ is the unit sphere. A point in this 5-volume is represented by the ray $(\xx, \nn)$ containing the location $\xx \in \RR ^3$ and the normalized approach vector $\nn \in \Ss ^2$ that lies on the unit sphere. We define the function $Shape_{\text{end}} : \DD \times \QQ \rightarrow \RR ^3 \times \Ss ^2$ that maps the configuration of the arm under a design to the ray at its end effector.

We also define the indicator function $I : \DD \times \EE \times \RR ^3 \times \Ss ^2 \rightarrow \{0, 1\}$ that indicates whether a particular ray can be reached by the end effector via a collision-free path in some environment $\ee$ under a design $\dd$:
\begin{equation}
I(\dd, \ee, \xx, \nn) =
\begin{cases}
1 & \begin{aligned}[t]
     &\text{if } \exists \sigma_{\text{free}} ^{\dd, \ee}, \tau \in [0,1] \text{ s.t.} \\
     &Shape_{\text{end}}(\dd, \sigma_{\text{free}} ^{\dd, \ee}(\tau)) = (\xx, \nn)
    \end{aligned} \\
0 & \text{otherwise}
\end{cases}
\end{equation}
Note that this indicator function requires the existence of a collision-free path from a start configuration to the particular ray rather than just a single collision-free configuration that reaches the ray.

To quantify the reachability of a design, we integrate the indicator function over the goal volume and unit sphere:
\begin{equation}
    \RVDSA (\dd, \ee) = \frac{1}{|\GG|} \int _{\GG} \int _{\Ss ^2} I(\dd, \ee, \xx, \nn)  \, d\nn \, d^3\xx
\end{equation}

which integrates over the unit sphere and the goal volume to calculate a measure representing the overall reachability of the approach vectors across the goal volume.

We average over the set of environments:
\begin{equation}
\RVDSA (\dd) = \frac{1}{|\mathbf{E} |}\sum _ {\ee \in \mathbf{E}}\RVDSA (\dd, \ee)
\end{equation}

Our goal is to find the optimal design:
\begin{equation}
    \dd ^* = \operatorname*{arg\,max}_{\dd} \RVDSA (\dd)
\end{equation}

\section{METHOD}
The main challenges of evaluating the objective function for a particular design are the continuous nature of $\GG \times \Ss ^2$ and the lack of a closed-form analytical solution for the inverse kinematics problem. We therefore discretize the goal hypervolume and use a JT-RRT inspired motion planner to compute an estimate of the objective function.

\subsection{Discretized Objective Function}

We discretize the 5-volume $\GG \times \Ss ^2  \subset \RR ^3 \times \Ss ^2$ by discretizing $\GG$ and $\Ss ^2$ independently. We discretize the goal volume $\GG$ into the finite set of voxels $\textbf{G}$. To discretize the unit sphere, we use the method of partitioning the sphere into a triangular mesh. We use the regular icosahedron, recursively subdividing each face into four separate triangles and reprojecting the vertices onto the unit sphere~\cite{Leibrandt2023_Designing}. This process yields the finite set $\textbf{T}$ of equilateral triangles. 

The goal 5-volume $\GG \times \Ss ^ 2$ is now discretized into the finite set $\textbf{G} \times \textbf{T}$ with $|\textbf{G}||\textbf{T}|$ elements. The definition of the discretized version of the indicator function is:
$\hat{I}(\dd, \ee, \xx, \nn) = 1 
\Leftrightarrow \exists \sigma_{\text{free}} ^{\dd, \ee}, \tau \in [0,1], \xx ' \in \text{ voxel}(\xx), \nn ' \in \text{ triangle}(\nn)
\text{ s.t. }
Shape_{\text{end}}(\dd, \sigma_{\text{free}} ^{\dd, \ee}(\tau)) = (\xx ', \nn ')$ where the voxel function returns the set of points within the voxel and the triangle function returns the set of points on the surface of the unit sphere within the projected triangular region of the discretized unit sphere. This expresses the idea that we consider a goal voxel reached if the end effector is anywhere within the voxel, and we consider a triangle at that goal voxel reached if the unit vector of the end effector's approach vector lies within the triangular region projected onto the unit sphere. We can calculate the solid angle coverage of a single unit sphere after determining what percentage of its representative triangles are reached.

We can define the discretized version of our objective function as follows:
\begin{equation}
\widehat{\RVDSA}(\mathbf{d}) = \frac{1}{|\mathbf{E}||\mathbf{T}|} \sum _{\ee\in \mathbf{E}} \frac{1}{|\textbf{G}_\ee|} \sum _{\xx \in \mathbf{G}_\ee} \sum _{\nn \in \mathbf{T}} \hat{I}(\dd, \ee, \xx, \nn)
\end{equation}

Our task is to find the optimal design based on this discretized objective function:
\begin{equation}
\hat{\dd ^*} = \operatorname*{arg\,max}_{\dd} \widehat{\RVDSA}(\dd)
\end{equation}

\subsection{Motion Planning}
We use a motion planner to efficiently compute the objective score $\RVDSA(\dd)$ for a given design. We use the Parallel Probabilistic Roadmap (PPRM) algorithm from the Motion Planning Templates library~\cite{Ichnowski2019_Motion}, biased to explore the task-relevant regions of the configuration space. The overall algorithm is outlined in Algorithm~\ref{alg:objective}.

From an initial configuration of $\mathbf{q} _0 = \textbf{0}$, we must reach the large number of directional triangles representing $\GG \times \Ss ^2$. To compute this efficiently, we exploit the fact that we wish to reach all of $\Ss ^2$, but only at a small subset of the workspace in $\RR ^3$. Rather than solving the inverse kinematics problems of reaching all $|\textbf{G}||\textbf{T}|$ rays $(\xx, \nn) \in \textbf{G} \times \textbf{T}$, we can instead solve the inverse kinematics problems of reaching the $|\textbf{G}|$ goal voxels $\xx \in \textbf{G}$ and then leveraging the end effector Jacobian to cover the $|\textbf{T}|$ directions at the goal voxel.

\begin{figure}[t]
    \centering
    \begin{subfigure}[t]{0.32\linewidth}
        \includegraphics[width=\linewidth]{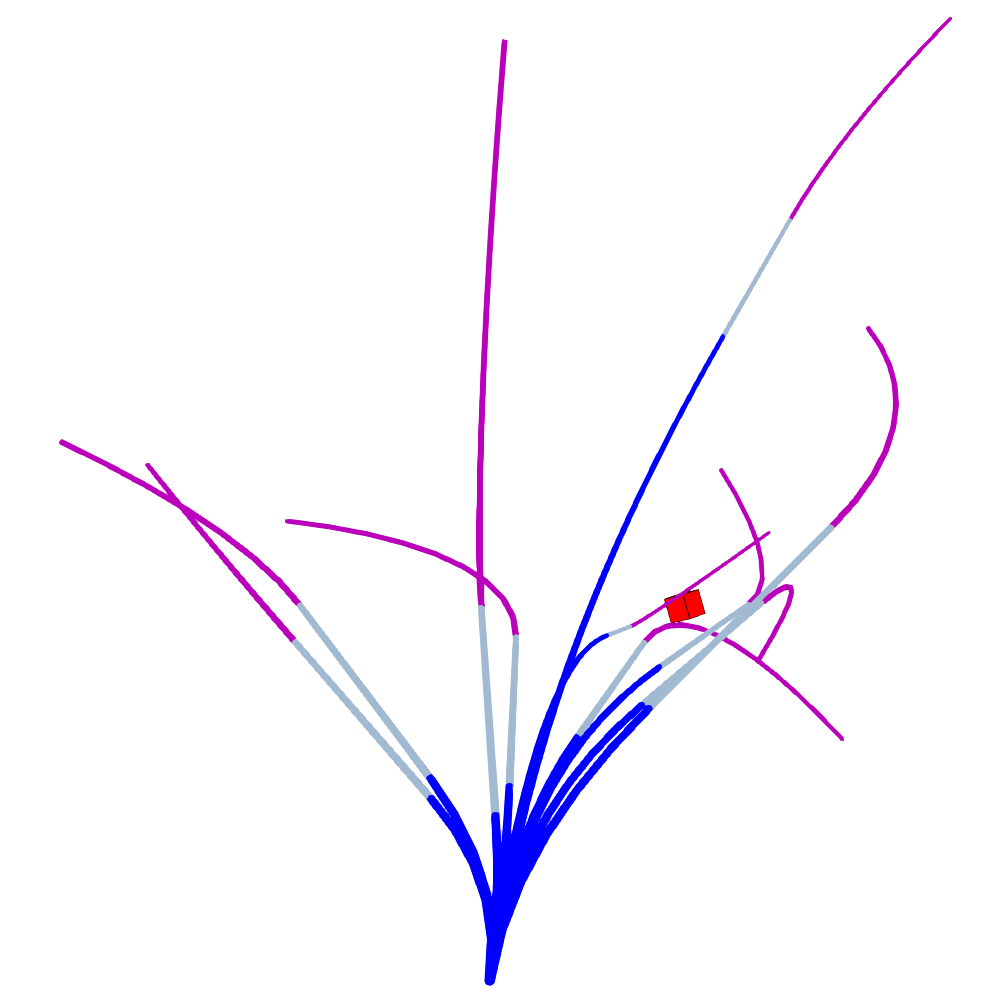}
        \caption{}
        \label{fig:subrand}
    \end{subfigure}
    \begin{subfigure}[t]{0.32\linewidth}
        \includegraphics[width=\linewidth]{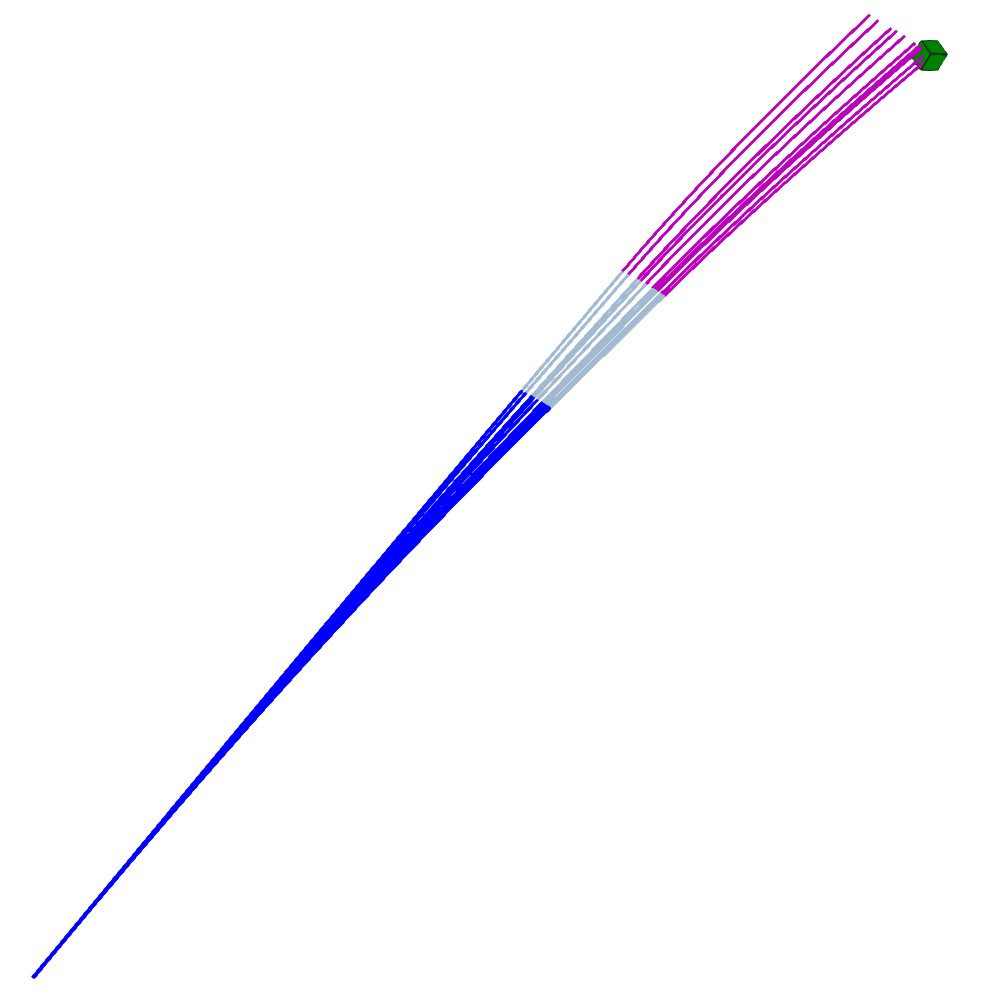}
        \caption{}
        \label{fig:subik}
    \end{subfigure}
    \begin{subfigure}[t]{0.32\linewidth}
        \includegraphics[width=\linewidth]{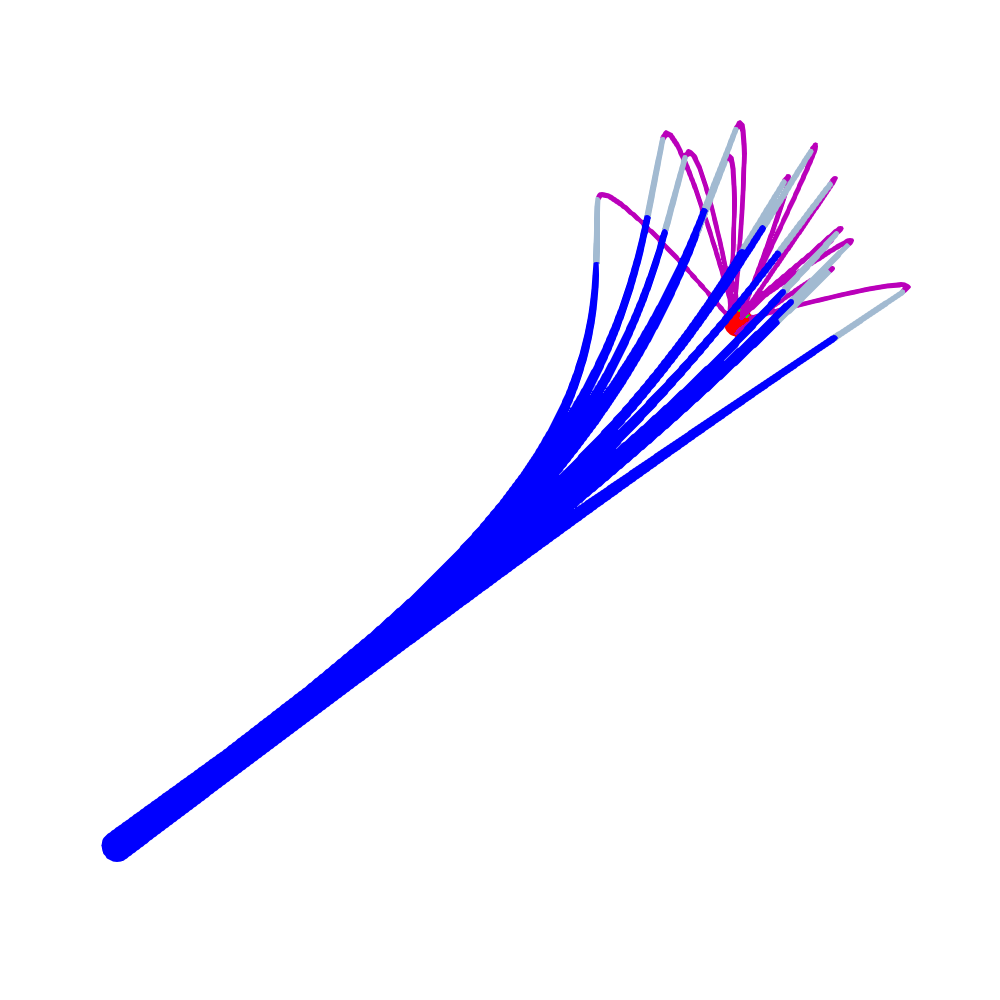}
        \caption{}
        \label{fig:subnull}
    \end{subfigure}
    
    \caption{(a) Random sampling. (b) Inverse kinematics sampling that brings the end effector to a goal voxel. (c) Self-motion sampling that reaches a goal voxel from many different directions.}
    \label{fig:coverage}
\end{figure}

To accomplish this, we choose a random goal voxel $\textbf{g} \in \textbf{G}$. Then, we query our PRM for the configuration for which the end effector is closest to $\textbf{g}$ using a nearest neighbors data structure \cite{Ichnowski2020_Concurrent} that is maintained simultaneously with the PRM. If the end effector location of this close configuration is not within the goal voxel $\textbf{g}$, then we use our sampling method of inverse kinematics sampling to find a new configuration that moves the end effector closer to the goal voxel. If the end effector is already within the goal voxel, then we use our sampling method of nullspace sampling to find different end effector directions at the goal voxel~\cite{Liegeois1977_Automatic}. The sampling methods are shown in Fig.~\ref{fig:coverage}.

\begin{algorithm}[t]
\caption{Objective Evaluation Function $\widehat{\RVDSA}(d)$}
\label{alg:objective}
\begin{algorithmic}[1]

\Require \hspace{0pt}
  \Statex \qquad $d$: design
  \Statex \qquad $O$: obstacles
  \Statex \qquad $G$: goal voxels
  \Statex \qquad $k$: node budget

\Ensure \hspace{0pt}
  \Statex \qquad $\widehat{\RVDSA}(d)$

\State $\text{PRM} \gets \Call{InitializePRM}{ }$
\State $\text{NodesEvaluated} \gets 0$
\While{\text{NodesEvaluated} $< k$}
  \State $p \gets \Call{RandomReal}{[0,1]}$
  \If{$p < \frac{1}{2}$}
    \State $q_{\text{new}} \gets \Call{RandomConfig}{d}$
  \Else
    \State $g \gets \Call{RandomGoalVoxel}{G}$
    \State $q_{\text{close}} \gets \Call{ClosestConfig}{\text{PRM},\, g}$
    \State $J \gets \Call{EndEffectorJacobian}{d, q_{\text{close}}}$
    \If{$\operatorname{Shape}_{\text{end}}(d, q_{\text{close}})$ is within voxel $g$}
      \State $\text{nullspace} \gets \Call{Nullspace}{J}$
      \State $\dot q _{\text{null}} \gets \Call{RandomVector}{\text{nullspace}}$
      \State $q_{\text{new}} \gets q_{\text{close}} + \alpha \cdot \dot q _{\text{null}}$
    \Else
      \State $(x, n) \gets \operatorname{Shape}_{\text{end}}(d, q_{\text{close}})$
      \State $\Delta x \gets g - x$
      \State $J^\ast \gets (J^T J + I)^{-1}\,J^T$
      \State $q_{\text{new}} \gets q_{\text{close}} + \alpha\, J^{\ast}\, \Delta x$
    \EndIf
  \EndIf
  \If{\Call{CollisionFree}{O, d, $q_{\text{new}}$}}
    \State \Call{AddSample}{\text{PRM}, $q_{\text{new}}$}
  \EndIf
  \State $\text{NodesEvaluated} \gets \text{NodesEvaluated} + 1$
\EndWhile
\State $\text{score} \gets \Call{CalculateRVDSA}{\text{PRM}}$
\State \Return score

\end{algorithmic}
\end{algorithm}

After the motion planner has run for its allotted node budget $k$, we can easily calculate the RVDSA by iterating through the nodes of the PRM that are in the same connected component as the start node and marking triangle faces as reached if the cached end effector pose reaches them.
\subsubsection{Random Sampling}
One strength of sampling-based motion planners is that the random sampling of the configuration space gives the property of probabilistic completeness. We retain this property of the PRM planner by sampling random configurations with a $50\%$ probability. While a different random sampling probability may improve performance, we choose $50\%$ as a simple heuristic.

\subsubsection{Inverse Kinematics Sampling}
Inspired by the JT-RRT method \cite{VandeWeghe2007_Randomized}, we sample configurations using a Jacobian-based controller in order to reach the goal voxels. Given some randomly selected goal voxel $\textbf{g} \in \textbf{G}$ that we wish to reach, the difference between the goal and the current end effector position is $\Delta \xx = \textbf{g} - \xx$.

The Jacobian of the end effector coordinates with respect to the configuration is defined by 
$\JJ (\textbf{q}) = \frac{\partial \xx}{\partial \qq}$ , hereby referred to as $\JJ$,
and we approximate this with the finite differences method. Since $\dot{\textbf{x}} = \JJ  \dot{\textbf{q}}$, we can use the pseudoinverse of the Jacobian to calculate the configuration space velocity to bring the end effector closer to the goal:
\begin{equation}
\qq _{\text{new}} = \qq + \alpha \JJ ^{*}  \Delta \xx
\end{equation}
where $\JJ ^*$ is the Damped Least Squares pseudoinverse \cite{Deo1995_Overview} and $\alpha = 0.01$ is a scaling factor.

\subsubsection{Self-Motion Sampling}
While the Jacobian-guided samples help the motion planner reach the 3D goal voxels, we use another strategy to increase coverage of the discretized unit sphere once the end effector has reached a goal voxel. Once some configuration leads to an end effector location within a goal voxel, we exploit this by exploring its self-motion manifold, configurations that keep the end effector position constant while varying its orientation.

Given a Jacobian matrix $\JJ$, its nullspace null$(\JJ)$ gives the set of configuration velocities which keep the 3D location of the end effector constant. We use singular value decomposition to obtain the basis vectors of the nullspace, and then we take a random linear combination of these vectors to obtain a velocity $\mathbf{\dot q} _{\text{null}}$. We calculate a new configuration
\begin{equation}
\qq_{\text{new}} = \qq + \alpha \mathbf{\dot q} _{\text{null}}
\end{equation}
where $\alpha = 0.01$ is a scaling factor.

This new configuration is on the self-motion manifold, keeping the end effector location nearly constant while changing its approach direction. This is efficient since it exploits the single configuration that reaches a goal voxel, quickly covering the range of possible approach vectors on the self-motion manifold.

For each generated configuration sample, we perform a collision check before adding it to our PRM. We precompute a 3D array by taking the Minkowski sum of the obstacle voxels and the largest radius of our dexterous sheaths tubes. By dilating the obstacles this way, we can efficiently check if the dexterous sheaths collide with the obstacles by checking along the centerline of the tubes.

\subsection{Design Optimization}
Our objective evaluation function $\widehat{\RVDSA}$ computes the score of any design $d$. We then use a global optimization algorithm to compute an optimized design. We use a modified version of Adaptive Simulated Annealing that increases the node budget $k$ of the motion planner at each design optimization iteration. Combining this with a design sampling function under which every neighborhood of the design space is sampled infinitely often provides asymptotic optimality \cite{Baykal2019_Asymptotically}.

\section{Application to Bimanual Dexterous Sheaths for Endoscopic Submucosal Dissection}
\begin{figure}[t]
\centering
\includegraphics[width=\linewidth]{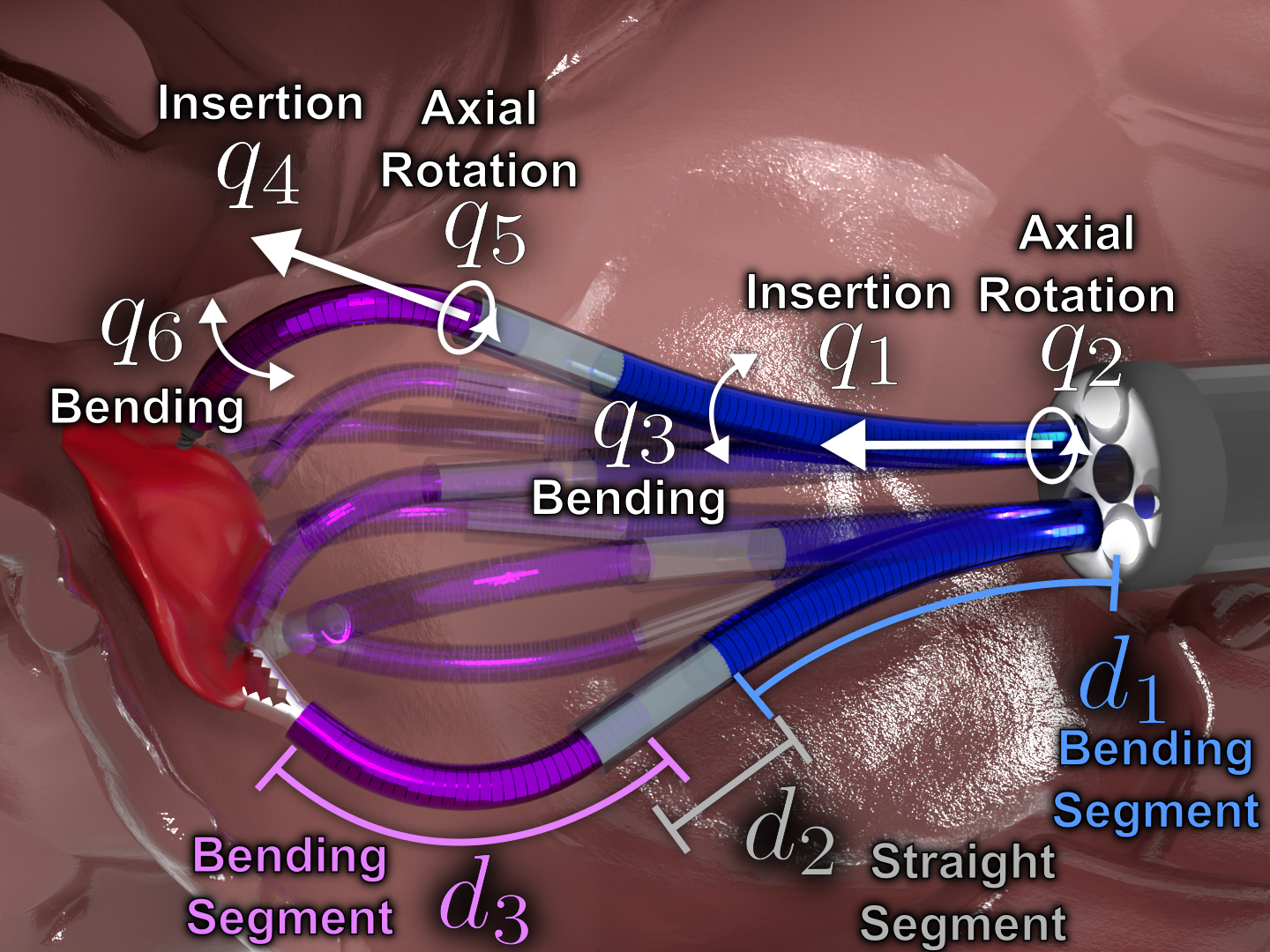}
\caption{Rendering of the bimanual dexterous sheaths robot in the colon reaching a cancerous polyp from many different directions. The design variables $d_1$ to $d_3$ are set at time of fabrication of the sheath's constituent tubes and remain fixed throughout a procedure. During a procedure, the robot is actuated through its degrees of freedom $q_1$ to $q_6$. Note that the tubes can be retracted so the full length of $d_1, d_2, d_3$ may not be exposed.}
\label{fig:robot}
\end{figure}

We apply our design optimization method to dexterous sheaths for Endoscopic Submucosal Dissection (ESD), a multi-stage procedure to remove large cancerous tumors in the colon. Dexterous sheaths are a type of continuum robotic manipulator consisting of thin, flexible arms that can operate in confined spaces \cite{Oliver-Butler2017_Concentric}. They use concentric push-pull tube pairs with custom-cut notch patterns and are connected at their tips. The tubes can controllably bend by pushing or pulling the pairs relative to each other~\cite{Oliver-Butler2017_Concentric}. They can be deployed through conventional endoscopes due to their small diameter and feature open lumens that allow endoscopic tools to pass through. Potential clinical applications include  tumor resection in airways \cite{Rox2018_Luminal} and colon cancer dissection \cite{Connor2025_System}.

While small polyps can be removed during a routine colonoscopy \cite{Williams2013_Management}, larger ones are more difficult to remove due to the limited dexterity of traditional endoscopic tools. Most people diagnosed with colorectal cancer in the US undergo a colectomy \cite{Wagle2025_Cancer}, in which the portion of the colon containing the cancerous polyp is removed. Therefore, we aim to lower the barrier to this minimally invasive ESD procedure by providing greater dexterity with the dexterous sheaths robot.
\subsection{Dexterous Sheaths Model}
We use constant curvature segments to model the kinematics of the tubes as this allows for efficient computation of forward kinematics and approximately models their kinematics \cite{Oliver-Butler2017_Concentric}. The design variables $d_1$ to $d_3$ combined with the configuration variables $q_1$ to $q_6$ specify three constant curvature segments representing the centerline of an arm.

\begin{table}[t]
\centering
\caption{Design Variables}
\label{tab:design}
\begin{tabular}{@{}lc@{}}
\hline
\textbf{Design Variable} & \textbf{Range} \\
\hline
$d_1:$ Outer pair length       & $[0,50]$ mm \\
$d_2:$ Outer pair straight section length  & $[0,50]$ mm \\
$d_3:$ Inner pair length         & $[0,50]$ mm \\
\hline
\end{tabular}
\end{table}

\begin{table}[]
\centering
\caption{Degrees of Freedom}
\begin{subtable}[t]{\linewidth}
\centering
\begin{tabular}{@{}lcc@{}}
\hline
\textbf{DoF} & \textbf{Space} & \textbf{Range} \\
\hline
$q_1:$ outer pair insertion       & $\mathbb{R}$ & $[0, 28]$ mm\\
$q_2:$ outer pair axial rotation  & $\mathbb{S}$ & $(-\infty, \infty)$ rad\\
$q_3:$ outer pair bending         & $\mathbb{R}$ & $[-0.0758, 0.0758]$ mm$^{-1}$\\
$q_4:$ inner pair insertion       & $\mathbb{R}$ & $[0,d_2]$ mm\\
$q_5:$ inner pair axial rotation  & $\mathbb{S}$ & $(-\infty, \infty)$ rad\\
$q_6:$ inner pair bending         & $\mathbb{R}$ & $[-0.1033, 0.1033]$ mm$^{-1}$\\
\hline
\end{tabular}
\caption{Left Arm DoF}
\end{subtable}

\vspace{1em}

\begin{subtable}[t]{\linewidth}
\centering
\begin{tabular}{@{}lcc@{}}
\hline
\textbf{DoF} & \textbf{Space} & \textbf{Range} \\
\hline
$q_1:$ outer pair insertion       & $\mathbb{R}$ & $[0, 28]$ mm\\
$q_2:$ outer pair axial rotation  & $\mathbb{S}$ & $(-\infty, \infty)$ rad\\
$q_3:$ outer pair bending         & $\mathbb{R}$ & $[-0.0880, 0.0880]$ mm$^{-1}$\\
$q_4:$ inner pair insertion       & $\mathbb{R}$ & $[0,d_2]$ mm\\
$q_5:$ inner pair axial rotation  & $\mathbb{S}$ & $(-\infty, \infty)$ rad\\
$q_6:$ inner pair bending         & $\mathbb{R}$ & $[-0.1225, 0.1225]$ mm$^{-1}$\\
\hline
\end{tabular}
\caption{Right Arm DoF}
\end{subtable}
\label{tab:dof}
\end{table}

\subsubsection{Design Space}

The variables of the optimization problem are the kinematic design parameters, which are optimized for a range of possible anatomies. The three design variables for our dexterous sheaths are listed in Table~\ref{tab:design} and shown in Fig.~\ref{fig:robot}. Each design variable applies to both arms.

\subsubsection{Configuration Space}
The dexterous sheaths robot has two arms, each with an outer pair and inner pair. Each tube pair can be inserted/retracted, rotated axially, and bent. Since there are two tube pairs per arm, each arm has $6$ degrees of freedom, as listed in Table~\ref{tab:dof} and shown in Fig.~\ref{fig:robot}. The two arms are identical except for the tube diameters, due to the two different working channel sizes of the Olympus CF-2T160I colonoscope.

There is a fundamental trade-off between higher max curvature (leading to higher dexterity) and device stiffness (leading to better manipulator effectiveness). Based on prior prototyping, we limit the curvature to $1/(4\cdot \text{tube diameter})$, and the values are shown in Table~\ref{tab:dof}.

The overlapping of bending sections can lead to an undesirable alignment of bending planes in these devices, which we call torsional snapping. Therefore, the outer pair can be retracted into the scope a maximum of 28 mm, past which the colonoscope itself can bend. We consider this fully retracted state $q_1 = 0$. Similarly, we limit the inner tube's retraction to the length of the straight segment, leading to a maximum insertion of $d_2$ for $q_4$.

\subsection{Colon Anatomy}
We use 5 colon anatomies with accompanying polyp data selected from the National CT Colonography Trial dataset~\cite{Smith2013_Data} and use 3D Slicer to create segmentations of the colon and polyp with the guidance of a radiology technologist, as shown in Fig.~\ref{fig:dataset}.

In the en bloc resection of a cancerous polyp, a lateral margin of around 5 mm is typically included around the polyp to ensure that no cancerous tissue is inadvertently left behind~\cite{Kawashima2023_Characteristics}. We use the Insight Toolkit (ITK) library \cite{McCormick2014_ITK} to compute the surface of the polyp with a 5 mm lateral margin to serve as the goal volume. We consider the colon voxels as obstacles for our motion planner since we do not wish for our dexterous sheaths to damage healthy tissue.

We choose voxels of size $1\,\text{mm} \times 1\,\text{mm} \times 1\,\text{mm}$ to voxelize the anatomy. For each goal voxel, we subdivide its associated icosahedron twice to produce a discretized $\Ss ^2$ consisting of $20 \cdot 4^2 = 320$ equilateral triangles. The average number of polyp voxels in each of the 5 anatomies is $668.2$, leading to $3341 \cdot 320 = 1,069,120$ triangles across the 5 anatomies and $2,138,240$ triangles across the two end effectors that are evaluated in each design optimization iteration.

\begin{figure}[t]
\centering
\includegraphics[width=\linewidth]{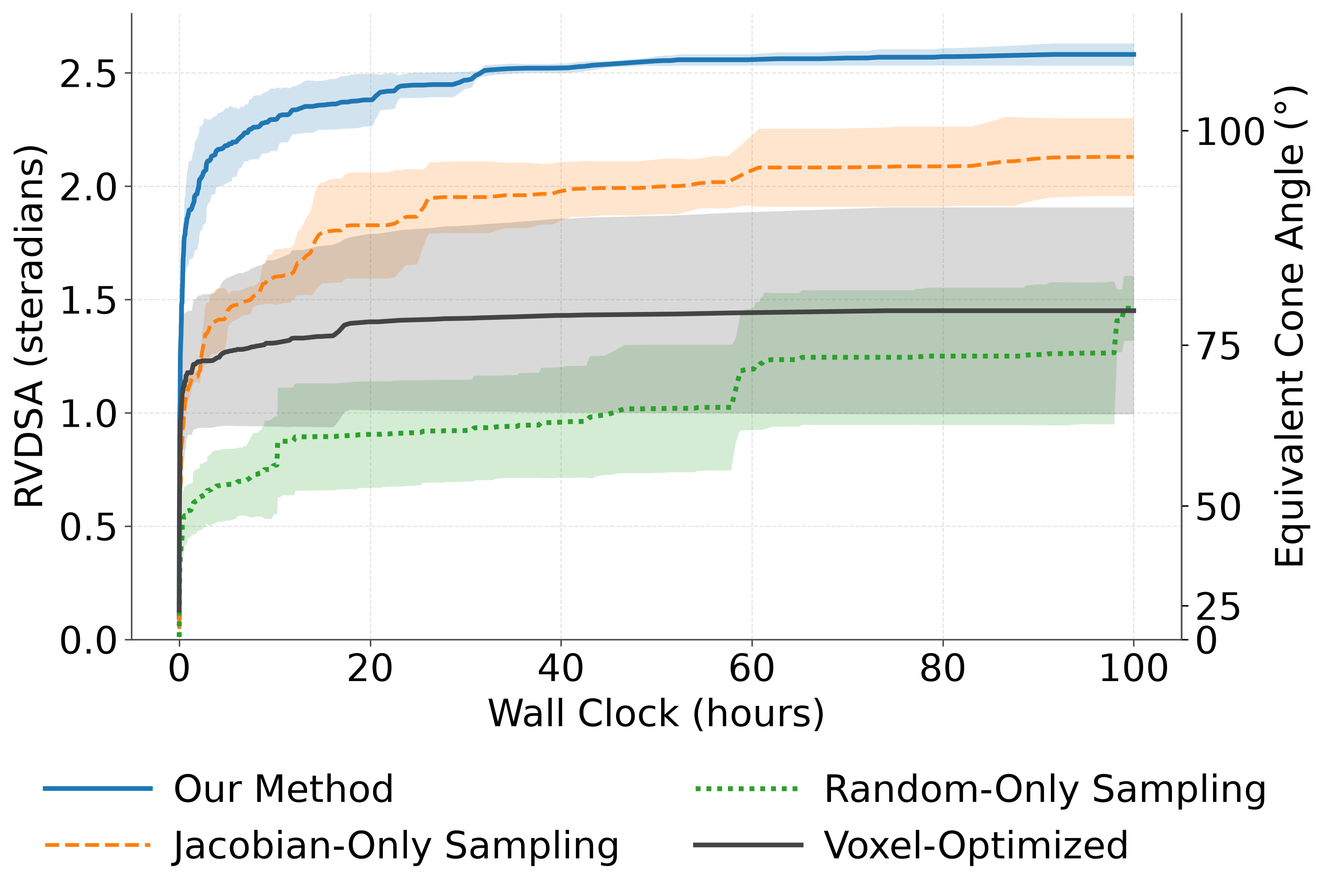}
\caption{The RVDSA is plotted as a function of run time for our method, Jacobian-only sampling, random-only sampling, and a voxel-optimized method. Each condition is run four times, with the mean and variance being displayed. The best RVDSA achieved by our method is $2.65$ steradians, which is equivalent to a cone angle of $109.3$ degrees.}
\label{fig:asa_results}
\end{figure}

We automatically select the start pose of the colonoscope. First, we compute the distance from the boundary for every voxel inside the colon \cite{Felzenszwalb2012_Distance} to then compute the centerline of the colon \cite{MingWan2002_Automatic}. We then choose a point on this centerline that has an unimpeded line of sight of the target volume. We exclude the centerline points that are past the polyp when traveling from the rectum. We choose the furthest point on the centerline, up to a distance of 50 mm (the middle of the scope's rated FOV depth), that has a line of sight on the center polyp voxel. With the location of the scope fixed, we orient the scope to point directly point toward the surface centroid. We perform this preprocessing step to compute the colonoscope's start pose for each of the 5 colon anatomies.

In our case with two end effectors, we average the RVDSA over both end effectors to obtain our objective function. This optimizes for the independent dexterity of each end effector, as this is a prerequisite for complex bimanual manipulation.

\section{RESULTS}
\begin{figure}[t!]
\centering
    \centering
    \begin{subfigure}[t]{\linewidth}
        \centering
        \includegraphics[width=0.49\linewidth]{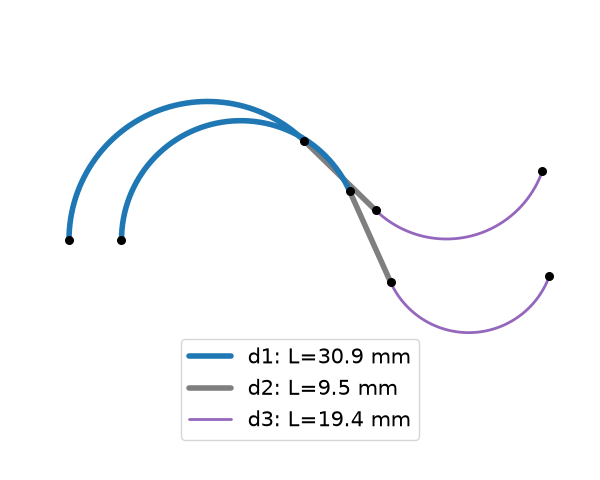}
        \includegraphics[width=0.49\linewidth]{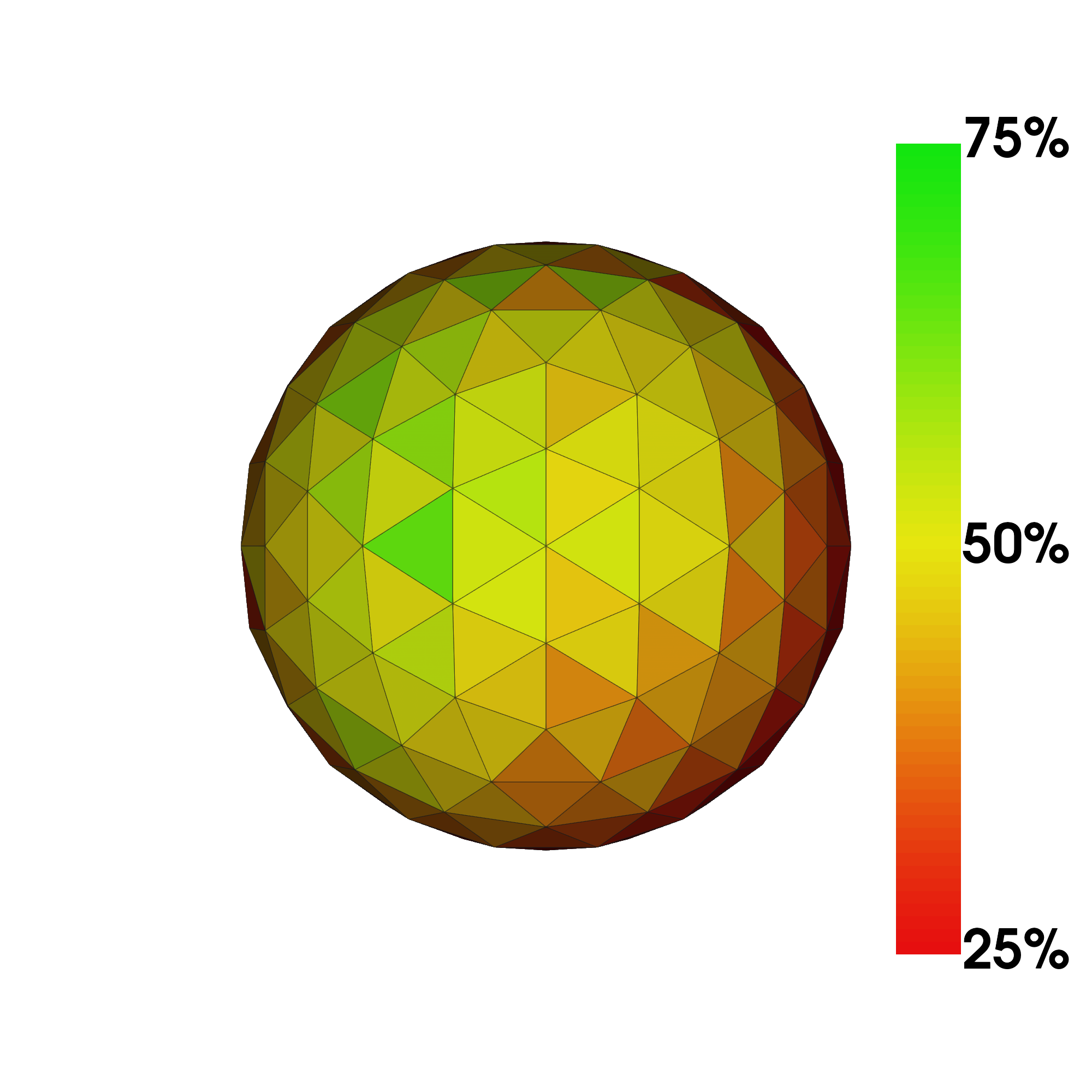}
        \caption{The best RVDSA-optimized design computed with our method during optimization across all runs. The RVDSA is $2.65$ steradians, or $109.3$ degree cone angle.}
    \end{subfigure}
    
    \begin{subfigure}[t]{\linewidth}
        \centering
        \includegraphics[width=0.49\linewidth]{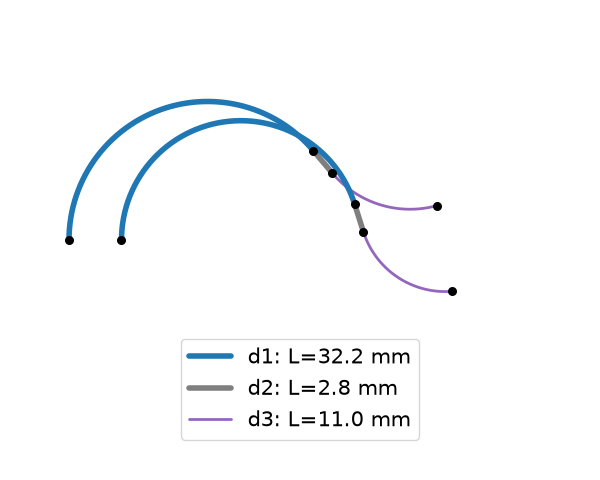}
        \includegraphics[width=0.49\linewidth]{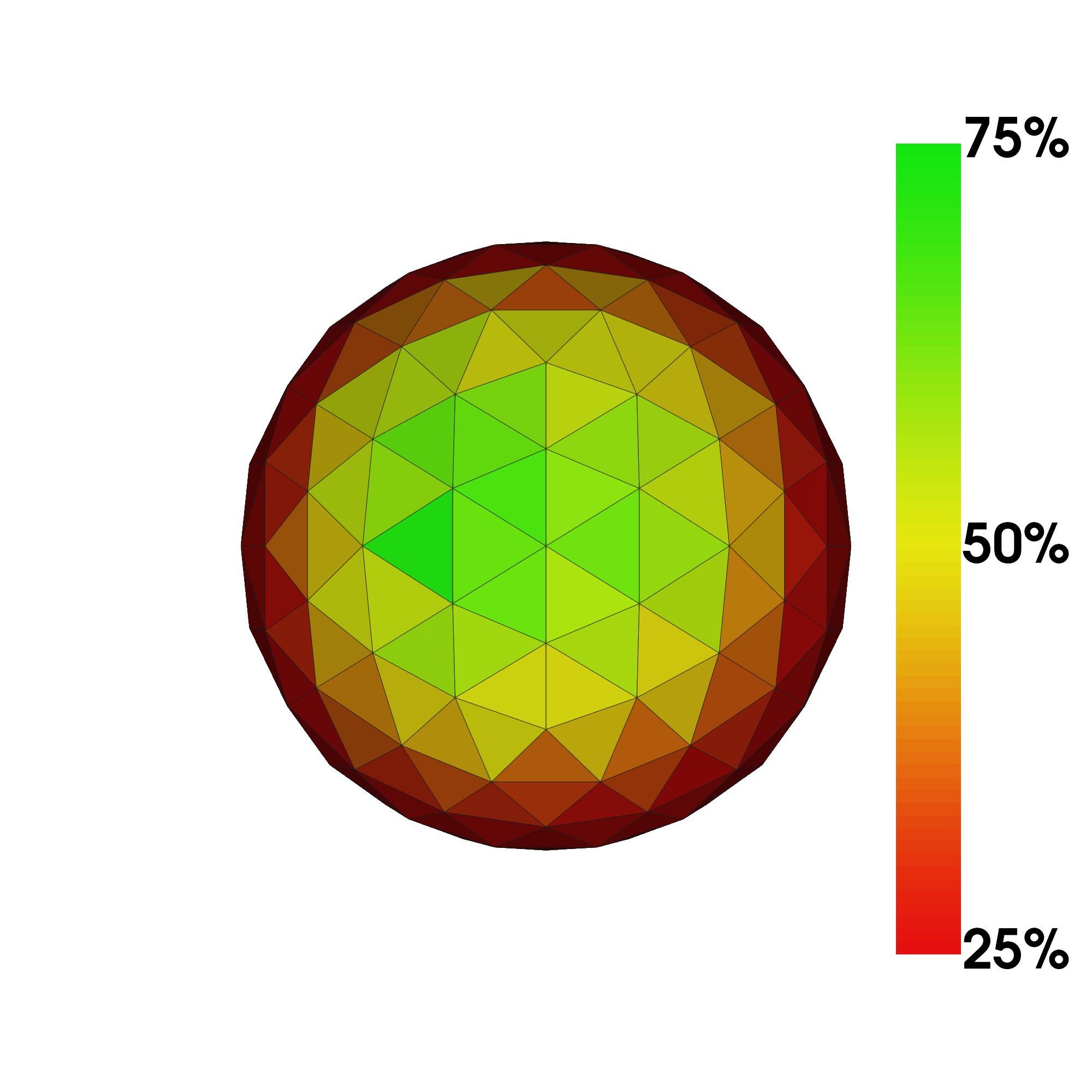}
        \caption{The best RVDSA-optimized design computed with random-only sampling during optimization across all runs. The RVDSA is $1.70$ steradians, or $86.4$ degree cone angle}
    \end{subfigure}

    \begin{subfigure}[t]{\linewidth}
        \centering
        \includegraphics[width=0.49\linewidth]{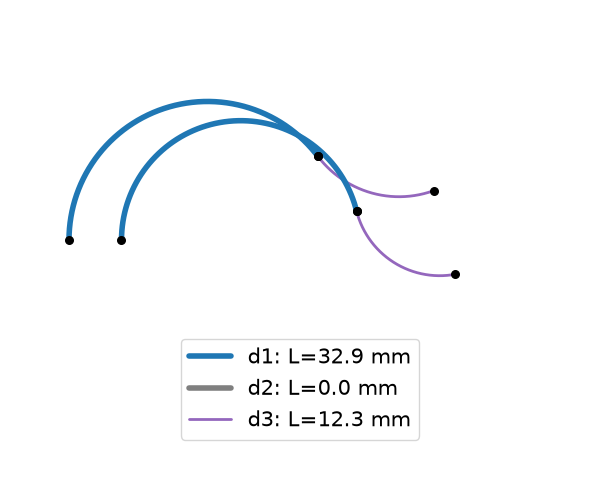}
        \includegraphics[width=0.49\linewidth]{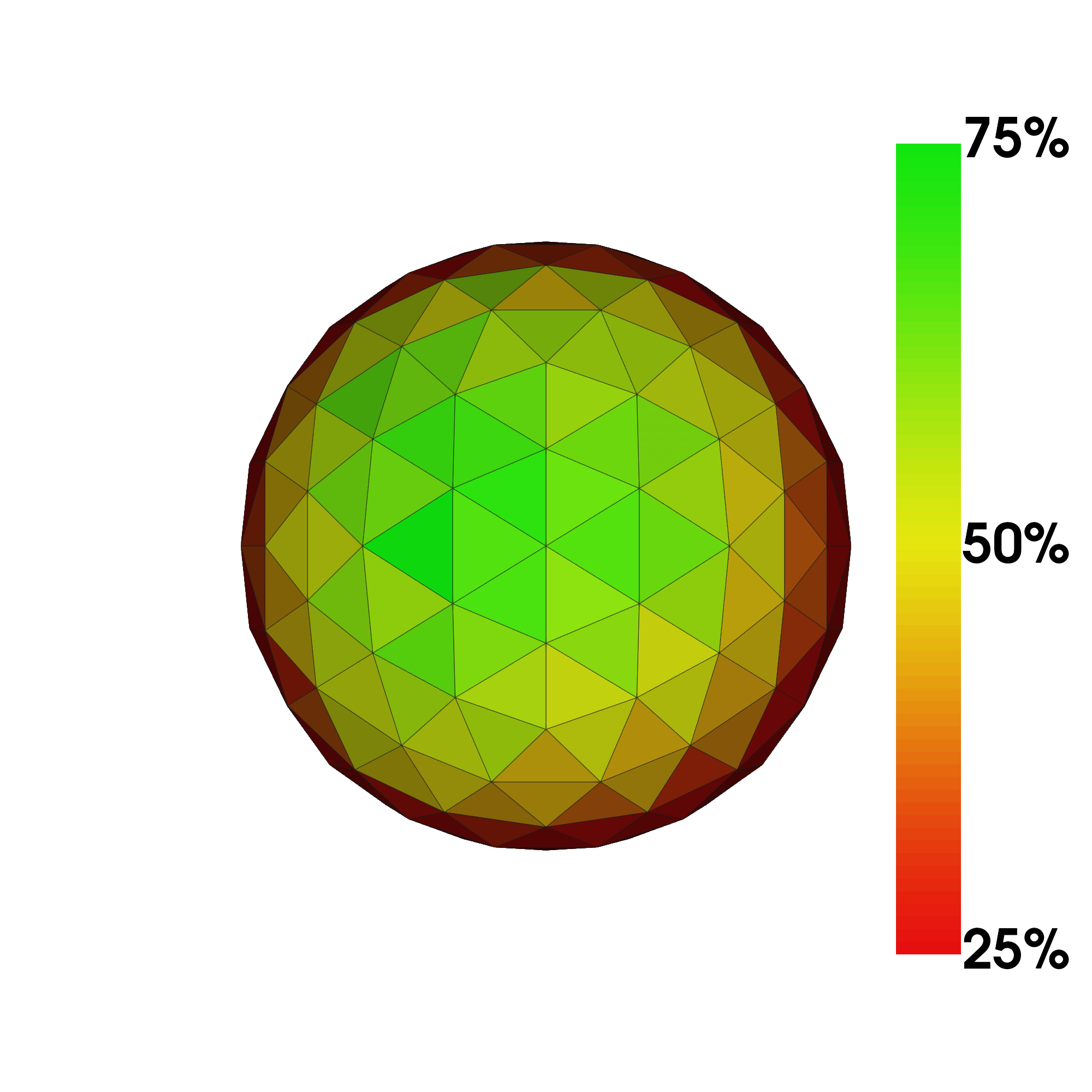}
        \caption{The best voxel-optimized design computed with our combined random and Jacobian sampling strategy during optimization across all runs. The RVDSA is $2.16$ steradians, or $98.0$ degree cone angle.}
    \end{subfigure}
    
    \caption{%The RVDSA-optimized design is compared against a voxel-optimized design across the 5 colon anatomies. 
    We compute the RVDSA of the best designs found across all runs for three different conditions. We display the discretized sphere coverage percentage for each individual triangle averaged over all polyp voxels and the two end effectors, after normalizing directions to the scope start frames.}
    \label{fig:designs}
\end{figure}

We ran our design optimization method with 5 colon anatomies on a 128-core CPU. In 100 hours of run time, our method ran approximately 4000 design iterations. We also ran design optimizations for three different conditions as comparisons. We ran our method with random-only sampling, which is the same as our method except without the Jacobian-guided sampling. We also ran our method with Jacobian-only sampling, which did not use any random sampling. Finally, we ran the voxel-optimized optimization method with the objective function replaced by the percentage of 3D goal voxels reached by the end effector. Each condition was run 4 times.

As shown in Fig.~\ref{fig:asa_results}, our method achieves the best results with an average RVDSA of $2.58$ steradians, with the best run resulting in an RVDSA of $2.65$ steradians. Our optimized design, shown in Fig.~\ref{fig:designs}, is able to reach goal voxels from more lateral directions. Our method, combining random sampling with inverse kinematics and self-motion sampling, performs better than either strategy separately. We attribute the poorer performance of the Jacobian-only sampling strategy to joint limits, singularities, and obstacles, which pure Jacobian control cannot escape.

The random-only sampling condition is similar to the method used by Leibrandt et al. \cite{Leibrandt2023_Designing} since it relies on randomly generated samples to cover the workspace. While random sampling will converge to the RVDSA given enough time, it does so much slower since configurations that reach $\GG \times \Ss ^2$ lie on low-dimensional manifolds, which random-only sampling reaches only rarely. This leads to an underestimation of the RVDSA within the motion planning budget. The optimizer therefore favors designs for which the RVDSA is easier to approximate under the budget, leading to an overall worse design, as shown in Fig.~\ref{fig:designs}. An accurate approximation of the RVDSA with random-only sampling would be computationally prohibitive given the thousands of designs evaluated in the design optimization loop.

We also include the performance of a voxel-optimized method in Fig.~\ref{fig:asa_results}. This is similar to prior work that only considered the coverage of a goal volume $\GG \subset \RR ^3$ \cite{Baykal2017_Asymptotically}. This method achieves a lower RVDSA since it is optimizing for coverage of the 3D voxels, as shown in Fig.~\ref{fig:designs}. The voxel-optimized design has a shorter straight segment $d_2$ and shorter distal curved segment $d_3$, which may be suited for voxel reachability but can only do so from limited directions.

The RVDSA-optimized design from our method has a longer straight segment $d_2$ and distal curved segment $d_3$ to enable reachability from different directions. Its ability to reach an equivalent cone angle of $109.3$ degrees across the polyp voxels in the five colon anatomies suggests that the computed design is highly dexterous and will likely be able to perform the required motions for ESD.

\section{CONCLUSION}

In this work, we presented a design optimization strategy that takes a surgical continuum robot's forward kinematics model and a set of anatomical environments, and returns a design that maximizes its Reachable Volumetric Dexterous Solid Angle on the provided environments. We use a motion planner with random and Jacobian-based sampling to efficiently compute the RVDSA without inverse kinematics, enabling design optimization over many iterations. We applied this to the dexterous sheaths robot to maximize its RVDSA for cancerous polyps in five colon anatomies. Our method achieves an average RVDSA of $2.58$ steradians, equivalent to an average cone angle of $107.8$ degrees. This is a $78\%$ improvement compared with the average RVDSA of designs optimized for 3D voxel coverage. Our optimized design will help determine the design of the next prototype of our dexterous sheaths system for Endoscopic Submucosal Dissection.

While optimizing for the independent dexterity of each arm does not account for cooperative tasks or collisions between the arms, having dexterous arms is a prerequisite for more complex interactions, which we will explore in future work. We will also increase the number of colon anatomies in the dataset by using automatic segmentation methods to improve the quality of the dataset and the resulting optimized design.

\section*{ACKNOWLEDGMENT}
We thank Elad Ohana for his assistance in preparing the colon anatomy dataset.

\bibliographystyle{IEEEtran}
\bibliography{refs}

\end{document}